\pdfoutput=1
\PassOptionsToPackage{prologue,dvipsnames}{xcolor}

\documentclass[11pt]{article}

\usepackage[]{ACL2023}

\usepackage{times}
\usepackage{latexsym}

\usepackage[T1]{fontenc}

\usepackage[utf8]{inputenc}

\usepackage{microtype}

\usepackage{inconsolata}

\usepackage{latexsym}
\usepackage{microtype}
\usepackage{lineno}
\usepackage{amsmath}
\usepackage{amssymb}
\usepackage{bbm}
\usepackage{multirow}
\usepackage{bm}
\usepackage{graphicx}
\usepackage{subfigure}
\usepackage[dvipsnames]{xcolor}
\graphicspath{{figures/}}

\title{DAPrompt: Deterministic Assumption Prompt Learning \\ for Event Causality Identification}

 \author{Wei Xiang \and Chuanhong Zhan \and Bang Wang \thanks{\quad Corresponding author: Bang Wang} \\
         School of Electronic Information and Communications, \\ 
         Huazhong University of Science and Technology, Wuhan, China \\
		\texttt{\{xiangwei, zhanch, wangbang\}@hust.edu.cn}} 

\begin{document}
\maketitle
\begin{abstract}
Event Causality Identification (ECI) aims at determining whether there is a causal relation between two event mentions. Conventional prompt learning designs a prompt template to first predict an answer word and then maps it to the final decision. Unlike conventional prompts, we argue that predicting an answer word may not be a necessary prerequisite for the ECI task. Instead, we can first make a deterministic assumption on the existence of causal relation between two events and then evaluate its rationality to either accept or reject the assumption. The design motivation is to try the most utilization of the encyclopedia-like knowledge embedded in a pre-trained language model. In light of such considerations, we propose a deterministic assumption prompt learning model, called DAPrompt, for the ECI task. In particular, we design a simple deterministic assumption template concatenating with the input event pair, which includes two masks as predicted events’ tokens. We use the probabilities of predicted events to evaluate the assumption rationality for the final event causality decision. Experiments on the EventStoryLine corpus and Causal-TimeBank corpus validate our design objective in terms of significant performance improvements over the state-of-the-art algorithms.
\end{abstract}

\section{Introduction}
\begin{figure}
	\centerline{\includegraphics[width=0.99\columnwidth, height=0.35\textwidth]{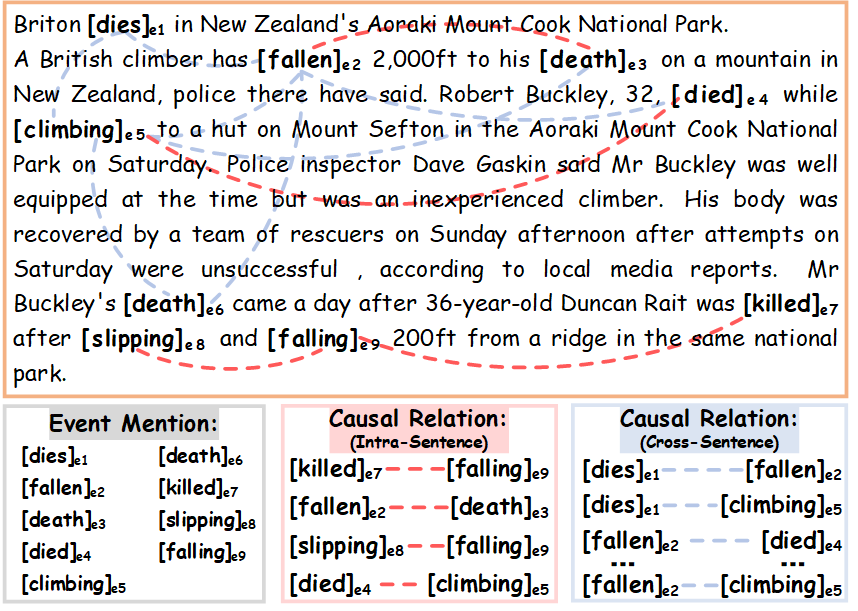}}
	\caption{Illustration of event causality annotation for an accident topic document in the ESC corpus. Event mentions are annotated as one or more words in a raw sentence, and causal relation annotations can exist in intra-sentence or cross-sentence event mentions.}
	\label{Fig:Example}
\end{figure}

Event Causality Identification (ECI) is to detect whether there exists a causal relation between two event mentions in a document. Fig.~\ref{Fig:Example} illustrates an example of event mention and causality annotations in an accident topic document in the widely used Event StoryLine Corpus (ESC), in which eleven event pairs are annotated with causal relation, including both the intra-sentence and cross-sentence causalities. The ECI task is to identify a causal relation between two event mentions. Causality identification is of great importance for many Natural Language Processing (NLP) applications, such as question answer~\cite{Bondarenko.A:et.al:2022:COLING,Sui.Y:et.al:2022:Knowl.BasedSyst.}, information extraction~\cite{Xiang.W:Wang.B:2019:Access}, and etc.

\par
Some recent deep learning-based methods design sophisticated neural models to learn a kind of contextual semantic representation for each event, such as the Rich Graph Convolutional Network (RichGCN)~\cite{Phu.M.T:Nguyen.T.H:2021:NAACL}, Event Relational Graph Transformer (ERGO)~\cite{Chen.M:et.al:2022:COLING}, Graph-based Event Structure Induction model (GESI)~\cite{Fan.C:et.al:2022:SIGIR}.
Although these graph neural networks can effectively learn contextual semantics as events' or event pairs' representations, they have ignored to utilize some external commonsense knowledge, like \textit{earthquake causes tsunami}, to augment causality detection.

\par
External knowledge bases can be employed to provide external causal knowledge for augmenting causality identification. For example, the \textit{ConceptNet}~\cite{Speer.R:et.al:2017:AAAI} contains abundant graph-structured knowledge, in which each node represents a concept and each edge corresponds to a semantic relation between concepts. Liu et al.~\cite{Liu.J:et.al:2020:IJCAI} and Cao et al.~\cite{Cao.P:et.al:2021:ACL} both use such knowledge triplets in the ConceptNet to boost representation learning. Moreover, the \textit{FrameNet} knowledge base~\cite{Baker.C.F:et.al:1998:ACL}, as well as the \textit{WordNet}~\cite{Miller.G.A:1995:Comm.ACM} and \textit{VerbNet}~\cite{Schuler.K.K:2006:PhD} lexical knowledge base have also been used to obtain external causal knowledge for the ECI task~\cite{Zuo.X:et.al:2021:ACL-Fingdings,Zuo.X:et.al:2020:COLING}.

\par
Although external knowledge bases can provide abundant information, how to extract appropriate knowledge triplets for the ECI task is not easy to implement, not to mention their encoding and fusion into task-specific events' representations. Recently, the \textit{pre-train, prompt, and predict} paradigm~\cite{Liu.P:et.al:2021:arXiv} (viz., \textit{prompt learning}) based on a Pre-trained Language Model (PLM) has been successfully applied in many NLP tasks. The successes can be contributed to the task transformation via carefully designed templates and answers, so as to well utilizing the encyclopedic linguistic and event causal knowledge embedded within a PLM during model training.

\par
For the ECI task, the basic idea of prompt learning is to design a prompt template such as \texttt{(...<$e_1$>[MASK]<$e_2$>...)}, and an answer space such as $\{ \texttt{cause}, \texttt{result from}, \texttt{not caused}, ...\}$. The template as a sentence is input into a PLM to output the mask token representation for its classification to an answer word, which is then further mapped into a causal relation. The recent DPJL model~\cite{Shen.S:et.al:2022:COLING} designs such a prompt template together with two derivative templates to augment representation learning for the mask token, which just achieved the new state-of-the-art performance of the intra-sentence event causality identification on the commonly used ESC corpus~\cite{Caselli.T:Vossen.P:2017:ACL-Ws}.

\par
We argue that the performance of such a conventional prompt learning is heavily dependent on the designed prompt templates and selected answer words. On the one hand, the manually-designed templates could be sensitive to its consisting words, as even synonyms (especially nouns/adj. words) could have subtle semantic differences that may impact on template quality. So a good template might need to try different combinations of composing synonyms. On the other hand, it is still a kind of \textit{implicit inference task} that transforms the ECI task into the prediction and mapping of some preselected answer word in the PLM vocabulary. As each answer word may also be kind of synonyms and with subtle semantic differences, it is often a big workload for selecting answers.

\par
Unlike conventional prompts, we argue that predicting an answer word may not be a necessary prerequisite for the ECI task. Instead, we can first assume that a causal relation does exist between two input events and then evaluate the rationality of such an assumption by directly predicting the input events. As such, we do not need to search for a well-designed prompt template as well as carefully selected answer words. Furthermore, predicting the input events from the raw sentences could better utilize a PLM for its powerful capability of learning contextual semantic representations, as well as utilizing some encyclopedic linguistic and event causal knowledge embedded within a PLM.

\par
Motivated from such considerations, we propose a novel \textit{deterministic assumption prompt learning} model, called DAPrompt, for the ECI task. Specifically, we first design a simple deterministic assumption template which includes two mask tokens for predicting the input events. We concatenate the two raw event sentences and the assumption template as an input sentence into a PLM. The objective is to predict the input events via the two masks for evaluating the rationality of the deterministic causal assumption. If the likelihood of correctly predicting the input events is larger than a decision threshold, then we accept the assumption and identify the existence of a causal relation. Experiment results show that our proposed DAPrompt significantly outperforms the state-of-the-art algorithms, in terms of much higher F1 score in all intra-sentence, cross-sentence, and overall event causality identifications\footnote{Source codes will be released after the anonymous review.}.

\section{Related Work}

\subsection{Graph-based Causality Identification}
The graph-based approaches first construct a graph and model the ECI as either a graph-based node classification or edge prediction problem.

\par
Some have applied graph neural networks for learning event node representations from document-level contextual semantics~\cite{Phu.M.T:Nguyen.T.H:2021:NAACL,Cao.P:et.al:2021:ACL,Fan.C:et.al:2022:SIGIR}.
For example, Phu and Nguyen~\cite{Phu.M.T:Nguyen.T.H:2021:NAACL} models diverse connections in between words of a document, like positional connection, syntactic dependency and etc., for the graph construction. They use a graph convolutional network to learn the event mention nodes' representations, and identify causalities through event node pair classification. Fan et al.~\cite{Fan.C:et.al:2022:SIGIR} build an event co-reference and co-occurrence graph to identify event causal relation with a graph convolutional network.

\par
Instead of node classification, some studies formalize the ECI task as a graph-based edge prediction problem~\cite{Zhao.K:et.al:2021:Inf.Sci.,Chen.M:et.al:2022:COLING}.
For example, Zhao et al.~\cite{Zhao.K:et.al:2021:Inf.Sci.} initialize event nodes' embeddings from a document-level encoder based on the PLM, and use a graph inference mechanism to update the graph for causal edge prediction.
Chen et al.~\cite{Chen.M:et.al:2022:COLING} build an event relational graph where each node denotes a pair of events and propose a graph transformer model to capture potential causal chains among nodes. These approaches, however, only exploit contextual and semantic information in a document for causal relation classification.

\subsection{Knowledge-boosted Causality Identification}
Some knowledge bases, like FrameNet~\cite{Baker.C.F:et.al:1998:ACL}, ConceptNet~\cite{Speer.R:et.al:2017:AAAI}, WordNet~\cite{Miller.G.A:1995:Comm.ACM}, VerbNet~\cite{Schuler.K.K:2006:PhD}, and etc., have been exploited to provide some kind of abstract causal knowledge.
\par
As those existing knowledge bases store a large amount of structured information, some studies directly exploit them for data expansion and augmentation in model training~\cite{Zuo.X:et.al:2020:COLING,Zuo.X:et.al:2021:ACL}. For example, Zuo et al.~\cite{Zuo.X:et.al:2020:COLING} extract the synonyms of each annotated causal event pair from lexical knowledge bases to distantly label event causality sentence as training data.
They further leverage existing knowledge bases to interactively generate new causal sentences for event causality data augmentation~\cite{Zuo.X:et.al:2021:ACL}.

\par
Besides data expansion, some studies try to discover the causal patterns from external knowledge bases to implement a kind of knowledgeable event causality inference~\cite{Liu.J:et.al:2020:IJCAI,Zuo.X:et.al:2021:ACL-Fingdings,Cao.P:et.al:2021:ACL}.
For example, Liu et al.~\cite{Liu.J:et.al:2020:IJCAI} propose to mine a kind of event-agnostic and  context-specific patterns from the ConceptNet to enhance the ability of their model for previously unseen cases.
Cao et al.~\cite{Cao.P:et.al:2021:ACL} encode some graph-structured knowledge from the ConceptNet, including descriptive graph knowledge and relational path knowledge, and performs event causality reasoning based on these induced knowledge. Zuo et al.~\cite{Zuo.X:et.al:2021:ACL-Fingdings} adopt a contrastive transfer learning framework to learn context-specific causal patterns from external causal statements.

\subsection{Prompt Learning Paradigm}
With the emergence of large-scale PLMs like the BERT~\cite{Devlin.J:et.al:2019:NAACL}, RoBERTa~\cite{Liu.Y:et.al:2019:arXiv}, and etc., the prompt learning has become a new paradigm for many NLP tasks, which uses the probability of text in PLMs to perform a prediction task and has achieved promising results~\cite{Seoh.R:et.al:2021:EMNLP,Wang.C:et.al:2021:EMNLP,Wei.X:et.al:2022:COLING}. For example, Seoh et al.~\cite{Seoh.R:et.al:2021:EMNLP} propose a cloze question prompt and a natural language inference prompt for aspect-based sentiment analysis.
Wang et al.~\cite{Wang.C:et.al:2021:EMNLP} propose a transferable prompting framework to capture cross-task knowledge for few-shot text classification.
Xiang et al.~\cite{Wei.X:et.al:2022:COLING} reformulate the implicit discourse relation recognition task as a connective-cloze prediction task and use prompt learning to predict an answer word and map it to a relation sense.

\par
A few studies have applied the prompt learning via designing appropriate prompt templates~\cite{Shen.S:et.al:2022:COLING,Liu.J:et.al:2020:IJCAI}. For example, Shen et al.~\cite{Shen.S:et.al:2022:COLING} use a masked language model as main prompt to predict the causality between event pair. They further design two derivative prompt task to leverage potential causal knowledge in PLM for explicit causality identification based on the causal cue word detection. Liu et al.~\cite{Liu.J:et.al:2020:IJCAI} use an event mention masking generalization mechanism to encode some event causality patterns for causal relation reasoning.

\par
The proposed DAPrompt is also based on the prompt learning paradigm, but it designs a novel prompting style of first deterministic assumption and next rationality evaluation.

\begin{figure*}[ht]
	\centering
	\includegraphics[width=0.95\textwidth, height=0.3\textwidth]{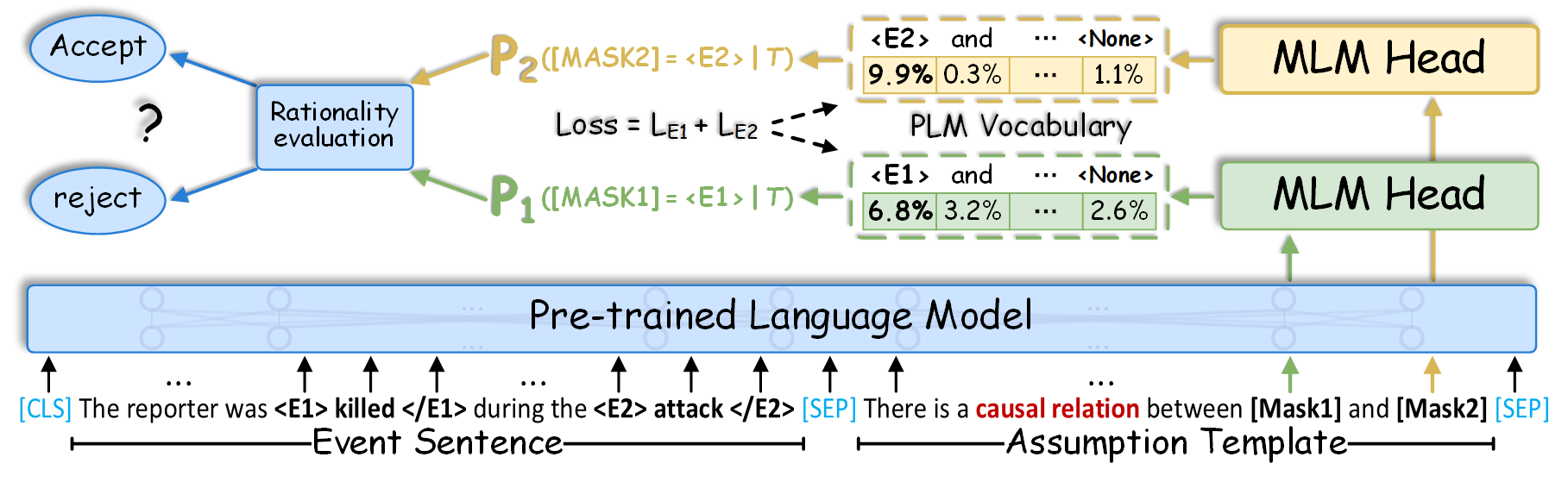}
	\caption{Illustration of our DAPrompt model.}
	\label{Fig:Model}
\end{figure*}

\section{The Proposed DAPrompt Model}
We first make a deterministic assumption on the existence of causal relation between two events in a document. Our DAPrompt identifies event causality by evaluating the rationality of a deterministic assumption. Specifically, we design a prompt template for a deterministic assumption to predict two input events, and use the probabilities of the correctly predicted events to determine whether to accept or reject the assumption, so as to making a final decision on event causality. Fig.~\ref{Fig:Model} illustrates the proposed DAPrompt model.

\subsection{Prompt Templatize}
The full prompt template $T$ contains two constructed sentences $T_1$ and $T_2$ that are concatenated with a \texttt{[SEP]} token as the input sentence to a PLM.

\par
The $T_1$, called the \textit{event sentence}, is designed for predicting two \textit{virtual event tokens} (VETs) \texttt{<E1>} and \texttt{<E2>}, each representing one of the input events. The design consideration is from the fact that the event mentions in different raw sentences usually consist of much different vocabulary words, not to mention having different lengths. We need to simplify and regulate their representations. We admit that using only two virtual tokens to represent diverse events is a bold attempt. Yet it provides an efficient way to link input events with the mask tokens in our assumption template.

\par
We note that although event mentions are normally annotated by a few words of a raw sentence, their representation learning should include the full sentence for better capturing contextual semantics. Let $S_1=(v_1, ...,[e_1,...,e_m],...,v_n)$ denote one raw sentence containing the annotated event mention $[e_1,...,e_m]$, where $v_i$s and $e_j$s are all vocabulary words. We insert the VET \texttt{<E1>} and \texttt{</E1>} before and after $[e_1,...,e_m]$ respectively to transform a raw sentence. The event sentence $T_1$ consists of a prefix token \texttt{[CLS]} and the two transformed sentences. Note that if two event mentions are within one raw sentence, we directly insert the VET tokens into the raw sentence to construct $T_1$. Fig.~\ref{Fig:Model} illustrates such an example of $T_1$ with one raw sentence containing two event mentions.

\par
The $T_2$ is our \textit{assumption template}, which designs a deterministic statement of the causal relation between two mask tokens, that is,
\begin{equation*}
	\small
	T_2 = \textrm{There is a {\color{BrickRed}{causal relation}} between} \texttt{[MASK1]} \textrm{and} \texttt{[MASK2]}
\end{equation*}
The two mark tokens are used to respectively predict the virtual event tokens. Let $\mathcal{V}$ denote the PLM vocabulary. The mask token \texttt{[MASK1]} is used to predict a word from $\mathcal{V}'_1=\mathcal{V}\cup \{E_1\}$, and the \texttt{[MASK2]} is used to predict a word from $\mathcal{V}'_2=\mathcal{V}\cup \{E_2\}$. Recall that a virtual event token is used to represent one event mention. So if both mask tokens can be correctly predicted as the corresponding virtual event token, then the deterministic causal assumption can be accepted, that is, there does exist a causal relation between the two events. The assumption template is suffixed with a separate \texttt{[SEP]} token.

\subsection{Answer Prediction}
We predict a mask token as one of the words in the enriched PLM vocabulary $\mathcal{V}'$. Two Masked Language Model (MLM) classifiers are adopted each for estimating the probability of a mask token as a vocabulary word:
\begin{align}
	P(\texttt{[MASK]} = v \in \mathcal{V}' \ | \  T).
\end{align}
Note that the two MLM classifiers are initially identical, which is pre-trained by the PLM. They will be separately fine-tuned during our model training. In each MLM classifier, a softmax layer is applied on the prediction scores of all words for the probability normalization.

\par
We are mainly interested in the following two probabilities:
\begin{equation}
	P_1 (\texttt{[MASK1]} = \texttt{<E1>} \ | \  T)\mathrm{\; , \;} P_2 (\texttt{[MASK2]} = \texttt{<E2>} \ | \  T).
\end{equation}
Each can be regarded as the likelihood of an input event appearing in the deterministic assumption template and will be used in our rationality evaluation.

\subsection{Rationality Evaluation}
We use the sum of $P_1$ and $P_2$ as a joint decision variable for \textit{rationality evaluation} of the deterministic assumption, that is,
\begin{align} \label{}
	f(T) = \begin{cases}
		Accept,	& \textit{if} \quad P_1 + P_2 \geq \rho \\
		Reject,	& \textit{if} \quad P_1 + P_2 \textless \rho
	\end{cases}
\end{align}
where $\rho$ is the \textit{joint decision threshold} and $\rho \in [0, 2]$ as $P_1, P_2 \in [0,1]$. If $P_1 + P_2 \geq \rho$, which suggests that the two masks are much likely to be the input events, then we accept the deterministic assumption of the existence of a causal relation between two events; Otherwise, we reject the assumption and the two input events are not with a causal relation. We note that we use a simple sum operation for $f(T)$, as we have no prior knowledge about which event is harder to predict.

\begin{table*}[htbp]
	\centering
	\resizebox{0.8\textwidth}{!}{
		\renewcommand\arraystretch{1.2}
		\begin{tabular}{l|ccc|ccc|ccc}
			\hline
			\multirow{2}{*}{Model}   & \multicolumn{3}{c|}{Intra-Sentence} & \multicolumn{3}{c|}{Cross-Sentence} & \multicolumn{3}{c}{Overall} \\ \cline{2-10}
			& P(\%)     & R(\%)     & F1(\%)     & P(\%)     & R(\%)     & F1(\%)     & P(\%)   & R(\%)   & F1(\%)  \\
			\hline
			ILP (NAACL, 2019)        & 38.8       & 52.4      & 44.6              & 35.1     & 48.2    & 40.6             & 36.2     & 49.5     & 41.9             \\
			RichGCN (NAACL, 2021)    & 49.2       & 63.0      & 55.2              & 39.2     & 45.7    & 42.2             & 42.6     & 51.3     & 46.6             \\
			GESI (SIGIR, 2022)       & -          & -         & 50.3              & -        & -       & \underline{49.3} & -        & -        & 49.4             \\
			ERGO (COLING, 2022)      & {57.5}     & {72.0}    & \underline{63.9}  & {51.6}   & 43.3    & 47.1             & 48.6     & 53.4     & \underline{50.9} \\ \hline
			{Our DAPrompt}           & {64.5}     & {73.6}    & \textbf{68.5}     & {59.9}   & {59.3}  & \textbf{59.0}    & {61.4}   & {63.7}   & \textbf{62.1}    \\
			\hline
	\end{tabular}}
	\caption{Overall results of comparison models for event causality identification on the EventStoryLine 0.9 corpus.}
	\label{Tab:ESC-Overall}
\end{table*}

\subsection{Training Strategy}
In the training phase, we use the \texttt{<E1>} and \texttt{<E2>} token as the positive label, if there is indeed a causal relation between two input events; While the virtual word \texttt{<None>} initialized by all other words is used as negative label for both \texttt{[MASK]} token prediction, if the causal relation assumption is incorrect.
We tune the PLM parameters as well as the two MLM classifier parameters based on these labels, and compute a cross entropy loss as a MLM classifier loss $\mathcal{L}_1$ ($\mathcal{L}_2$):
\begin{align} \label{}
	\mathcal{L} = -\frac{1}{K} \sum\limits_{k=1}^{K} \mathbf{y}^{(k)} \log(\mathbf{\hat{y}}^{(k)}) + \lambda \Vert \theta \Vert^2,
\end{align}
where $\mathbf{y}^{(k)}$ and $\mathbf{\hat{y}}^{(k)}$ are the answer label and predicted answer of the $k$-th training instance, respectively. $\lambda$ and $\theta$ are the regularization hyper-parameters. The overall loss of our DAPrompt is as follows:
\begin{align} \label{}
	\mathcal{L}_{DAPropmt} = \mathcal{L}_1 + \mathcal{L}_2.
\end{align}
We use the AdamW optimizer~\cite{Loshchilov.I:Hutter.F:2019:ICLR} with $L2$ regularization for model training.

\section{Experiments Settings}
In this section, we present our experiment settings, including the dataset, PLMs, competitors, and parameter settings.

\subsection{Datasets}
Our experiments are conducted on two widely used datasets for the ECI task, that is, the EventStoryLine 0.9 Corpus (ESC) ~\cite{Caselli.T:Vossen.P:2017:ACL-Ws} and the Causal-TimeBank Corpus (CTB)~\cite{Mirza.P:Tonelli.S:2014:COLING}. 

\par
\textbf{EventStoryLine} contains 22 topics and 258 documents from various news web-sites. There are in total 5,334 event mentions in the ECS dataset. A total number of 5,655 event pairs are annotated with causal relations, among which 1,770 causal relations are from intra-sentence event pairs and 3,855 causal relations are from cross-sentence event pairs. Following the standard data splitting~\cite{Gao.L:et.al:2019:NAACL}, we use the last two topics as the development set, and conduct 5-fold cross-validation on the remaining 20 topics. The average results of precision, recall, and F1 score are adopted as performance metrics.

\par
\textbf{Causal-TimeBank} contains 184 documents from English news articles and 7,608 annotated event pairs.
A total of 318 event pairs are annotated with causal relations, among which 300 causal relations are from intra-sentence event pairs and only 18 causal relations are from cross-sentence event pairs.
Following the standard data splitting~\cite{Liu.J:et.al:2020:IJCAI}, we employ a 10-fold cross-validation evaluation and the average results of precision, recall, and F1 score are adopted as performance metrics.
Following~\cite{Phu.M.T:Nguyen.T.H:2021:NAACL}, we only conduct intra-sentence event causality identification experiments on CTB, as the number of cross-sentence event causal pairs is quite small.

\par
\subsection{Competitors}
We compare our DAPrompt with the following competitors:
\begin{itemize}
	\item \textsf{KnowDis}~\cite{Zuo.X:et.al:2020:COLING} uses external lexicon knowledge to distantly augment event causality data.
	\item \textsf{KnowMMR}~\cite{Liu.J:et.al:2020:IJCAI} uses external knowledge to mine some event causality patterns.
	\item \textsf{CauSeRL}~\cite{Zuo.X:et.al:2021:ACL-Fingdings} adopts a contrastive strategy to transfer learned external causal statements.
	\item \textsf{LearnDA}~\cite{Zuo.X:et.al:2021:ACL} leverages existing knowledge bases to interactively generate training data.
	\item \textsf{LSIN}~\cite{Cao.P:et.al:2021:ACL} uses a graph induction model to learn external structural and relational knowledge.
	\item \textsf{DPJL}~\cite{Shen.S:et.al:2022:COLING} leverages two derivative prompt tasks to identify the explicit and implicit causality.
	\item \textsf{ILP}~\cite{Gao.L:et.al:2019:NAACL} uses integer linear programming to identify causal relations based on document-level causal structures constraints.
	\item \textsf{RichGCN}~\cite{Phu.M.T:Nguyen.T.H:2021:NAACL} uses a graph convolutional network to learn a document context-augmented representation of event pairs.
	\item \textsf{GESI}~\cite{Fan.C:et.al:2022:SIGIR} builds an event co-reference graph to identify event causal relation by a graph convolutional network.
	\item \textsf{ERGO}~\cite{Chen.M:et.al:2022:COLING} builds an event relational graph using an event pair as a node to capture causation transitivity via a transformer-like neural network.
\end{itemize}

\par
\subsection{Parameter Setting}
We implement the PLM models with their 768-dimension base version provided by the HuggingFace transformers\footnote{https://github.com/huggingface/transformers}~\cite{Wolf.T:et.al:2020:EMNLP}, and run PyTorch~\footnote{pytorch.org} framework with CUDA on NVIDIA GTX 3090 GPUs. We set the mini-batch size to 16, the learning rate to 1e-5, the determine threshold $\rho$ to 0.6, and all trainable parameters are randomly initialized from normal distributions.
As the positive and negative samples are unbalanced, we adopt a random negative sampling with probability of 0.2 on the training dataset.

\section{Results and Analysis}

\subsection{Overall Results}
Table~\ref{Tab:ESC-Overall} and Table~\ref{Tab:ESC-Intra} compare the overall performance between our \textsf{DAPrompt} and the competitors on the ESC corpus; While Table~\ref{Tab:CTB-Intra} compares the performance on the CTB corpus. The competitors in Table~\ref{Tab:ESC-Overall} have reported all intra-sentence, cross-sentence, and overall results on the ESC dataset; While the competitors in Table~\ref{Tab:ESC-Intra} and Table~\ref{Tab:CTB-Intra} have only reported the intra-sentence results on ESC and CTB datasets, respectively.
Besides, we note that the competitors in Table~\ref{Tab:ESC-Overall} all exploit some document-level information to encode event mentions for causality classification; While the competitors in Table~\ref{Tab:ESC-Intra} all use some kind of external knowledge to enhance event causality identification.

\begin{table}
	\centering
	\resizebox{0.9\columnwidth}{!}{
		\renewcommand\arraystretch{1.2}
		\begin{tabular}{l|ccc}
			\hline
			\multirow{2}{*}{Model} & \multicolumn{3}{c}{Intra-Sentence}  \\ \cline{2-4}
			& P(\%)   & R(\%)   & F1(\%)  \\ \hline
			KnowDis (COLING, 2020)       & 39.7      & 66.5      & 49.7       \\
			KnowMMR (IJCAI, 2020)        & 41.9      & 62.5      & 50.1       \\
			CauSeRL (ACL-Findings, 2021) & 41.9      & 69.0      & 52.1       \\
			LearnDA (ACL, 2021)          & 42.2      & 69.8      & 52.6       \\
			LSIN (ACL, 2021)             & 47.9      & 58.1      & 52.5       \\
			DPJL (COLING, 2022)          & 65.3      & 70.8      & 67.9       \\ \hline
			Our DAPrompt                 & \textbf{64.5}      & \textbf{73.6}      & \textbf{68.5}       \\ \hline
	\end{tabular}}
	\caption{Overall results of event causality identification on the EventStoryLine 0.9 corpus.}
	\label{Tab:ESC-Intra}
\end{table}

\par
The first observation is that the \textsf{RichGCN}, \textsf{GESI}, and \textsf{ERGO} can obviously outperform the \textsf{ILP} in Table~\ref{Tab:ESC-Overall}.
This might be attributed to the use of some graph-based neural networks, operating on the document-level graph structure with large-scale trainable parameters to augment event representation learning. Indeed, graph-based neural networks have been proven to be effective for many NLP tasks~\cite{Scarselli.F:et.al:2009:IEEETrans.NN,Piao.Y:et.al:2022:AAAI}. We can also observe that the improvement of intra-sentence causality identification is more significant than that of cross-sentence. This might be attributed to the use of pre-trained language model for event node encoding, which can capture the semantic interaction between two events in a sentence.

\begin{table}
	\centering
	\resizebox{0.9\columnwidth}{!}{
		\renewcommand\arraystretch{1.2}
		\begin{tabular}{l|ccc}
			\hline
			\multirow{2}{*}{Model} & \multicolumn{3}{c}{Intra-Sentence}  \\ \cline{2-4}
			& P(\%)   & R(\%)   & F1(\%)  \\ \hline	
			KnowMMR (IJCAI, 2020)        & 36.6 & 55.6 & 44.1 \\
			KnowDis (COLING, 2020)       & 42.3 & 60.5 & 49.8 \\
			RichGCN (NAACL, 2021)        & 39.7 & 56.5 & 46.7 \\
			LearnDA (ACL, 2021)          & 41.9 & 68.0 & 51.9 \\
			CauSeRL (ACL-Findings, 2021) & 43.6 & 68.1 & 53.2 \\
			LSIN (ACL, 2021)             & 51.5 & 56.2 & 53.7 \\
			ERGO (COLING, 2022)          & 62.1 & 61.3 & 61.7 \\
			DPJL (COLING, 2022)          & 63.6 & 66.7 & 64.6 \\ \hline
			Our DAPrompt                 & \textbf{66.3} & \textbf{67.1} & \textbf{65.9} \\ \hline
	\end{tabular}}
	\caption{Overall results of event causality identification on the Causal-TimeBank corpus.}
	\label{Tab:CTB-Intra}
\end{table}

\par
The second observation is that the \textsf{DPJL} adopting the prompt learning paradigm can significantly outperform the other competitors in Table~\ref{Tab:ESC-Intra}.
The outstanding performance can be attributed to the task transformation for directly predicting a PLM vocabulary word, other than fine-tuning a downstream task-specific neural model upon a PLM. Although these competitors have used some kind of extra knowledge, such as lexicon knowledge and relational knowledge, from large-scale external knowledge bases, the prompt learning model can better enjoy the encyclopedic linguistic knowledge embedded in a PLM during the model training.
Similar results can also be observed  on the CTB dataset in Table~\ref{Tab:CTB-Intra}.

\par
Finally, our \textsf{DAPrompt} (using DeBERTa as the PLM) has achieved significant performance improvements over all competitors in terms of much higher F1 score with all intra-sentence, cross-sentence, and overall event causality identification on both ESC and CTB datasets.
We attribute its outstanding performance to our task transformation of evaluating the rationality of a deterministic assumption: We do not need to predict an unknown relation between events, no matter what kind of relations could be. Instead, we only need to evaluate the causal rationality via a deterministic assumption between two input events. We note that even some competitors have achieved outstanding performance in the intra-sentence case, as presented in Table~\ref{Tab:ESC-Intra} and Table~\ref{Tab:CTB-Intra}, they cannot be applied in the cross-sentence case. For example, the \textsf{DPJL} has exploited a causal cue word for event causality identification, but such a cue word does not exist in a cross-sentence event pair.

\begin{figure}[t]
	\centering
	\subfigure[\label{SubFig:individual} Individual thresholds (Event-1/Event-2)] 
	{\includegraphics[width=0.45\textwidth]{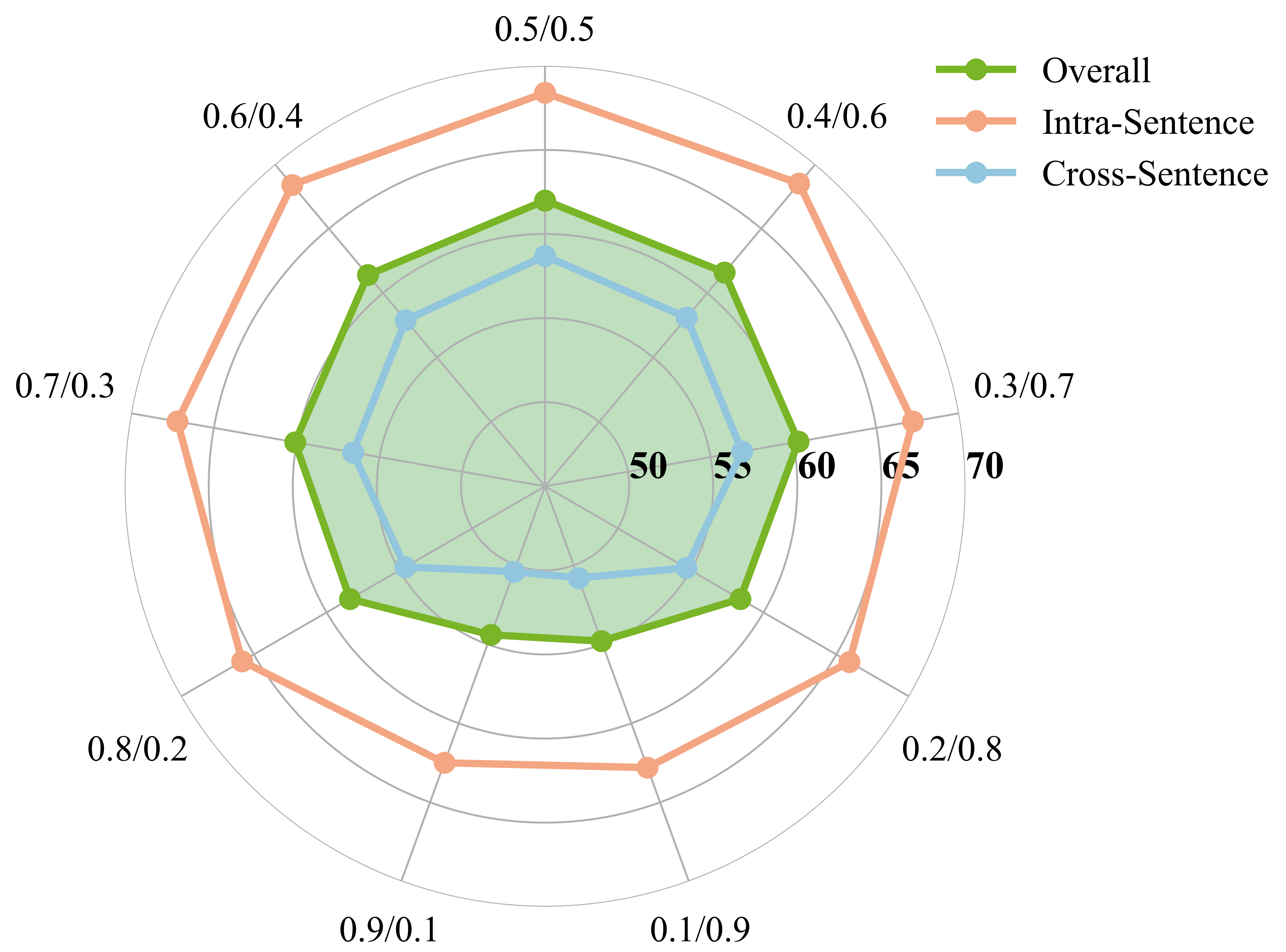}} 
	\subfigure[\label{SubFig:Joint} Joint thresholds] 
	{\includegraphics[width=0.42\textwidth,height=0.18\textwidth]{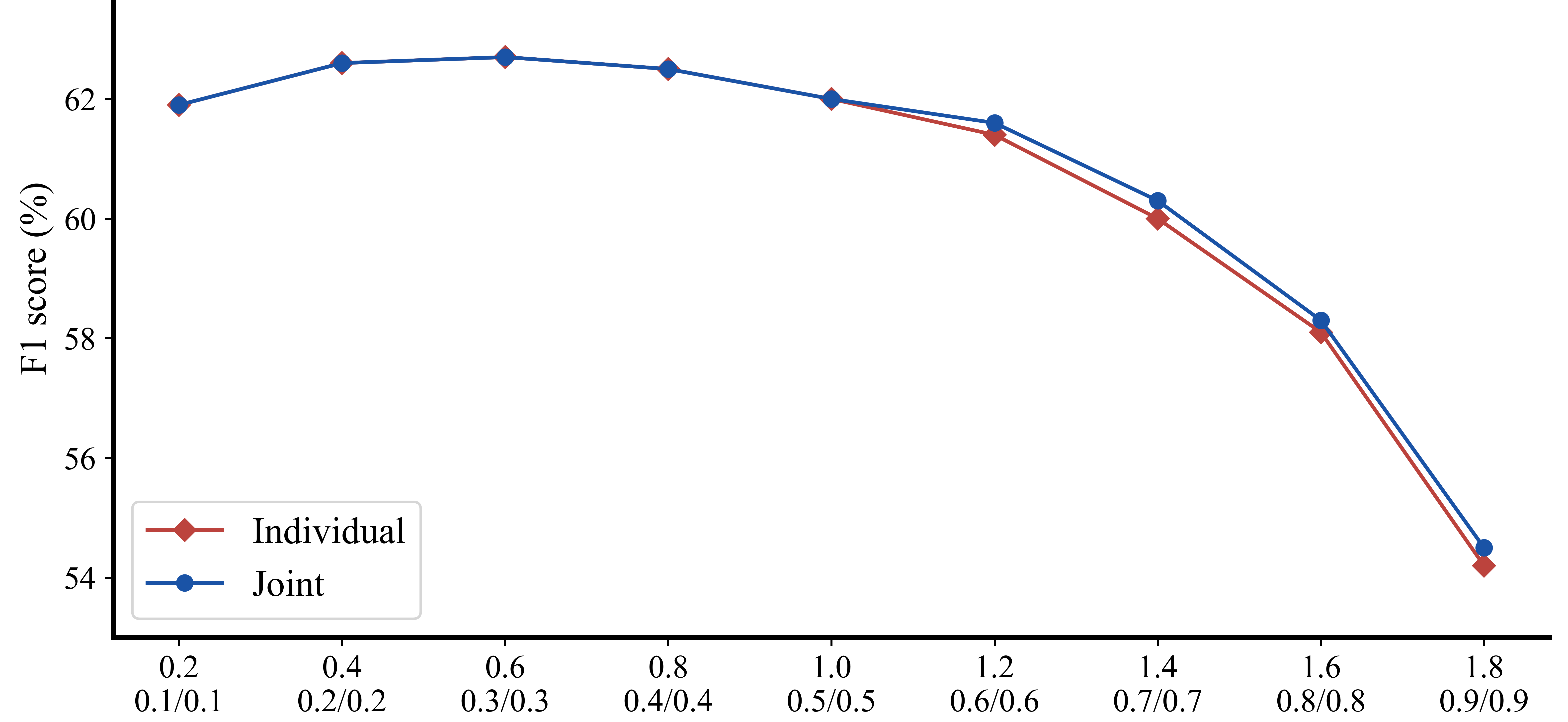}} 
	\caption{Performance comparison between using individual threshold and joint threshold on the ESC corpus.}
	\label{Fig:Threshold-Setting}
\end{figure}

\begin{figure*}[h]
	\centering
	\hspace{-3mm}
	\subfigure[\label{SubFig:F1-Score} F1 Score]
	{\includegraphics[width=0.35\textwidth]{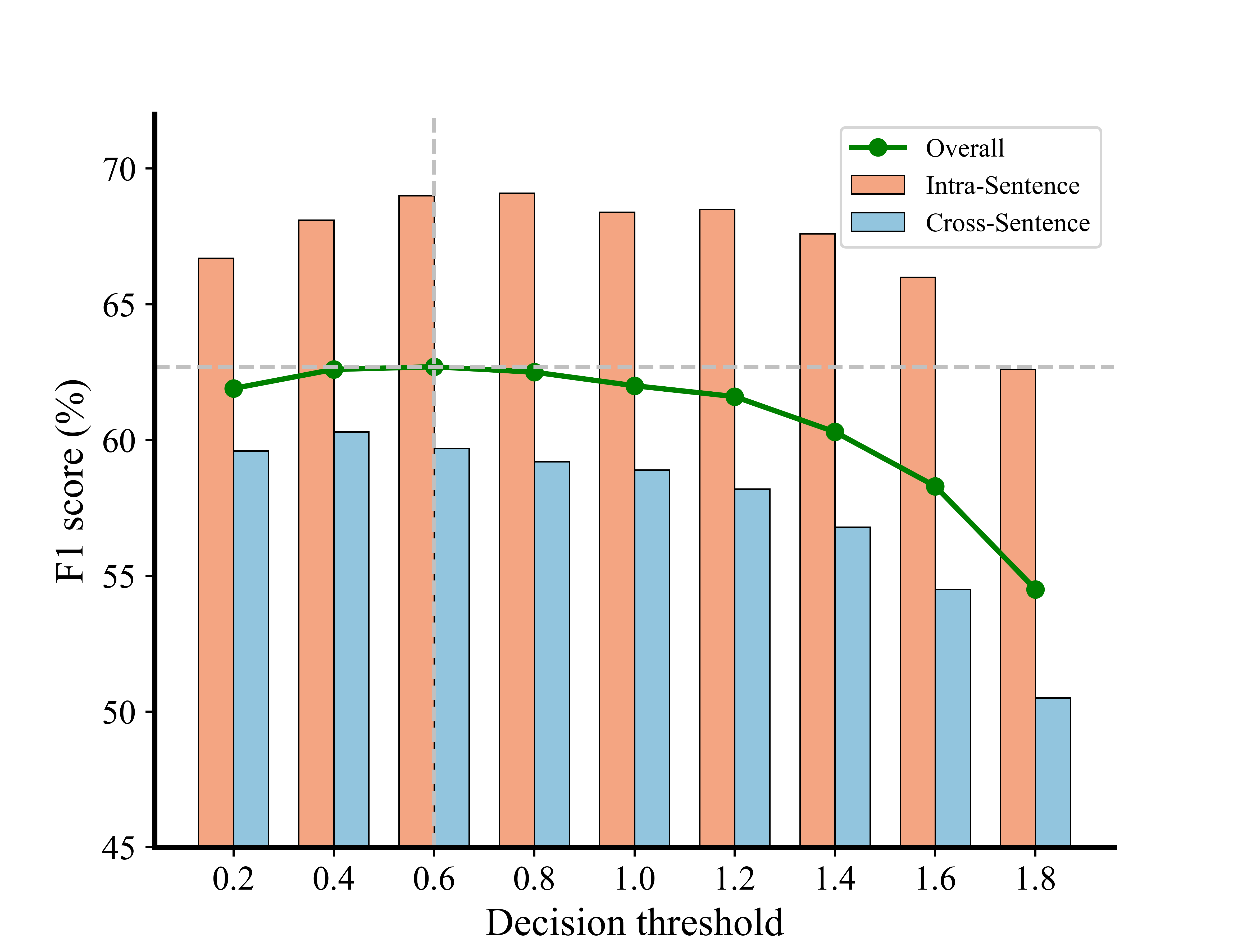}} \hspace{-5mm}
	\subfigure[\label{SubFig:Precision} Precision]
	{\includegraphics[width=0.35\textwidth]{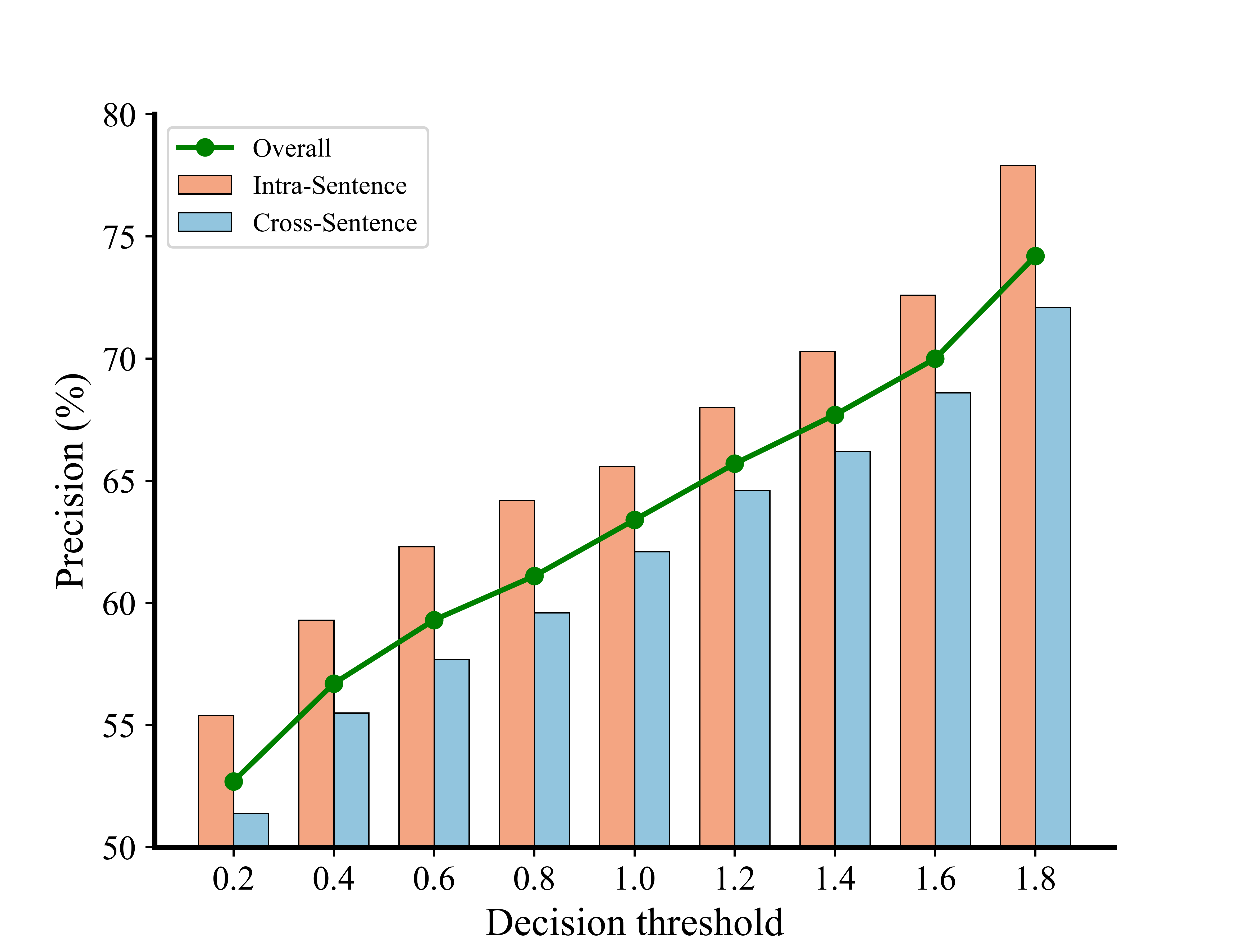}} \hspace{-5mm}
	\subfigure[\label{SubFig:Recall} Recall]
	{\includegraphics[width=0.35\textwidth]{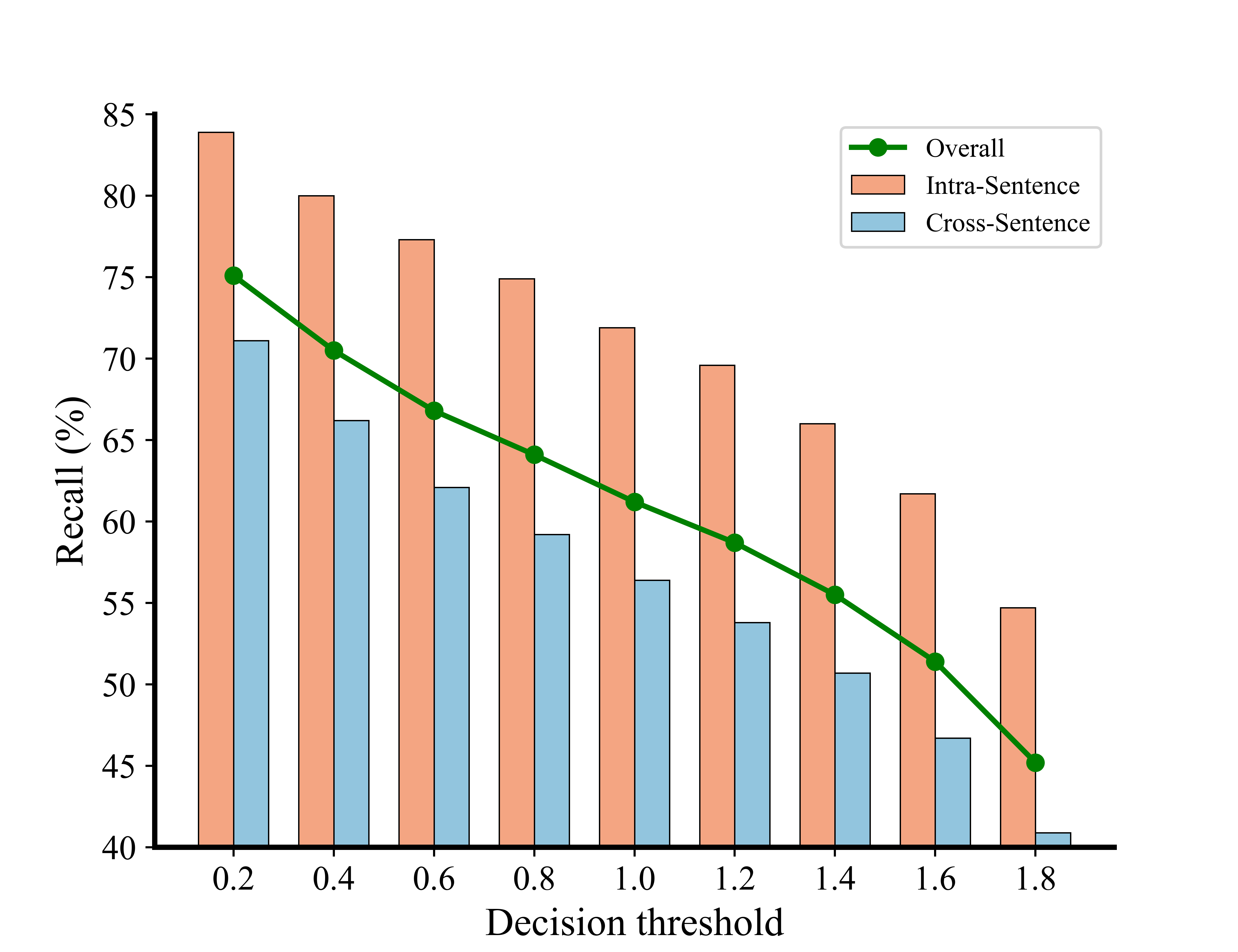}} \hspace{-5mm}
	\caption{Performance comparison of using different decision thresholds on the ESC corpus.}
	\label{Fig:Threshold}
\end{figure*}

\par
\textbf{Decision Threshold:} To examine the effectiveness of different decision threshold strategies, we conduct experiments on both individual threshold and joint threshold with different threshold values.
The joint threshold strategy is that we use a joint decision variable $P_1+P_2$ and a joint decision threshold $\rho$. The individual threshold strategy is that we use two \textit{individual decision thresholds} $\rho_1$ and $\rho_2$ for $P_1$ and $P_2$, respectively. If $P_1 \geq \rho_1$ and $P_2 \geq \rho_2$, we accept the deterministic assumption that a causal relation exists between two events.

\par
Fig.~\ref{Fig:Threshold-Setting} (a) plots the performance of our \textsf{DAPrompt} (DeBERTa) using individual decision threshold in rationality evaluation on the ESC corpus. Each corner of the radar map represents a decision threshold ratio for two events, and the closer a point to the corner, the better performance of identifying event causality.
It can be observed that \textsf{DAPrompt} achieves the best performance when the discrimination threshold is set equally for both events, i.e. (0.5/0.5); While \textsf{DAPrompt} suffers from an imbalance discrimination threshold setting, such as (0.1/0.9), (0.9/0.1), and etc. This indicates that the rationality of both events may be significant for identifying the causal relation between them. As we have no prior knowledge about the importance of each event, we simply sum their probabilities for rationality evaluation.

\par
Fig.~\ref{Fig:Threshold-Setting} (b) compares the overall performance of \textsf{DAPrompt} (DeBERTa) between using equal individual decision threshold and the joint decision threshold on the ESC corpus. It can be observed that \textsf{DAPrompt} achieves nearly the same F1 score within a large range of the decision threshold (the range [0.2, 1.0] in the figure) using these two kinds of decision threshold settings. Yet the performance of \textsf{DAPrompt} using equal individual decision threshold cannot outperform the joint decision threshold when the decision threshold is set in the range of [1.0, 1.8].
This can be attributed to the flexibility of using a joint decision threshold, allowing two events to be identified as having a causal relation, even if one event has slightly lower rationality but the other event has higher rationality.
For such considerations, we adopt the joint decision threshold in our \textsf{DAPrompt}.

\par
Fig.~\ref{Fig:Threshold} plots the performance of our \textsf{DAPrompt} (DeBERTa) against using different joint decision thresholds in rationality evaluation on the ESC corpus. We can observe that our \textsf{DAPrompt} achieves the best overall performance in terms of the F1 score when the discrimination threshold is set to 0.6. Yet the overall performance does not change much within a large range of the decision threshold (the range $[0.2,1.0]$ in the figure). This suggests the wide applicability of our model for its not much sensitive to the decision threshold.

\par
We can also observe from Fig.~\ref{Fig:Threshold} that our \textsf{DAPrompt} suffers from either a very large or very small value of the discrimination threshold. Indeed, a small decision threshold relaxes the requirement for correctly predicting the input events, which thus admits too many assumed causal relations to be accepted. As such, the recall is high yet the precision is small. By contrast, a large decision threshold tightens the event prediction requirement, which only allows those event predictions with high confidence to accept a deterministic assumption. As such, the precision is high yet the recall is small. From our experiments, we suggest to take an empirical setting around 0.6 for the decision threshold.

\begin{table}
	\centering
	\resizebox{0.8\columnwidth}{!}{
		\renewcommand\arraystretch{1.2}
		\begin{tabular}{l||c|c|c}
			\hline
			Model  & Intra- & Cross- & Overall \\ \hline
			DAPrompt (BERT)     & 68.1      & 58.5      & 61.6       \\
			DAPrompt (RoBERTa)  & 68.3      & 58.1      & 61.3       \\
			DAPrompt (ERNIE)    & 68.1      & 56.9      & 60.7       \\
			DAPrompt (DeBERTa)  & 68.5      & 59.0      & 62.1       \\
			\hline
	\end{tabular}}
	\caption{Experiment results of using different PLM (F1 score).}
	\label{Tab:PLM}
\end{table}

\subsection{Ablation Study}
\textbf{Pre-trained Language Model:} In the prompt learning, using different PLMs may impact on the task performance. Table~\ref{Tab:PLM} compares the results of our proposed \textsf{DAPrompt} on the ESC corpus adopting the most representative PLMs:
\begin{itemize}
	\item \textsf{BERT}~\cite{Devlin.J:et.al:2019:NAACL}: The most representative PLM proposed by Google~\footnote{https://github.com/google-research/bert}, which is pre-trained using a \textit{cloze task} and a \textit{next sentence prediction} task.
	\item \textsf{RoBERTa}~\cite{Liu.Y:et.al:2019:arXiv}: A BERT-enhanced PLM proposed by Facebook~\footnote{https://github.com/pytorch/fairseq/}, which removes the next sentence prediction objective and is pre-trained on a much larger dataset with some modified key hyper-parameters.
	\item \textsf{ERNIE}~\cite{Sun.Y:et.al:2019:arXiv}:	A knowledge-enhanced PLM proposed by Baidu~\footnote{https://github.com/PaddlePaddle/ERNIE}, which uses some knowledgeable masking strategies in pre-training.
	\item \textsf{DeBERTa}~\cite{He.P:et.al:2021:ICLR}: The latest masked PLM proposed by Microsoft~\footnote{https://github.com/microsoft/DeBERTa}, which improves BERT and RoBERTa models using a disentangled attention mechanism and an enhanced mask decoder.
\end{itemize}

\begin{table*}[t]
	\centering
	\resizebox{0.90\textwidth}{!}{
		\renewcommand\arraystretch{1.2}
		\begin{tabular}{l||ccc||ccc|ccc|ccc}
			\multicolumn{1}{c}{} & \multicolumn{3}{c}{\textbf{Causal-TimeBank}}  & \multicolumn{9}{c}{\textbf{EventStoryLine}} \\
			\hline
			\multicolumn{1}{c||}{\multirow{2}{*}{Model}} & \multicolumn{3}{|c||}{Intra-Sentence} & \multicolumn{3}{c|}{Intra-Sentence} & \multicolumn{3}{c|}{Cross-Sentence} & \multicolumn{3}{c}{Overall}  \\
			\multicolumn{1}{c||}{}                       & P(\%)     & R(\%)     & F1(\%)     & P(\%)     & R(\%)     & F1(\%)     & P(\%)   & R(\%)   & F1(\%)  & P(\%)      & R(\%)     & F1(\%)     \\
			\hline
			Prompt                 & 52.1 & 51.4 & 51.2     & 63.9  & 66.8  & 65.1  & 52.9  & 46.0   & 48.9   & 56.7  & 52.5  & 54.2      \\
			Prompt + VA            & 58.7 & 51.7 & 54.2     & 59.9  & 73.2  & 65.7  & 49.9  & 52.3   & 50.3   & 53.3  & 58.8  & 55.3       \\
			Prompt + CT            & 58.2 & 51.7 & 53.4     & 61.7  & 69.8  & 64.8  & 53.3  & 50.7   & 49.7   & 56.1  & 56.7  & 54.7       \\
			Prompt + VA + CT   & 55.9 & 56.4 & 55.9     & 62.0  & 70.3  & 65.5  & 52.5  & 50.8   & 51.0   & 55.7  & 56.9  & 55.8       \\
			\hline
			DAPrompt w/ SiM    & 60.7 & 57.1 & 58.6     & 56.6  & 56.3  & 55.7  & 57.3  & 54.8   & 55.7   & 57.3  & 55.1  & 55.7       \\
			DAPrompt w/ ShM   & 64.6 & 59.1 & 61.3     & 59.4  & 75.1  & 66.0  & 56.2  & 65.1   & 59.5   & 57.3  & 68.2  & 61.6        \\
			DAPrompt w/ ET      & 22.3 & 12.1 & 14.6     & 60.5  & 42.7  & 49.6  & 39.2  & 38.6   & 38.6   & 44.3  & 39.9  & 41.7        \\
			DAPrompt full (ours)                & 66.3 & 67.1 & \textbf{65.9}     & 64.5  & 73.6  & \textbf{68.5}  & 59.9  & 59.3   & \textbf{59.0}   & 61.4  & 63.7  & \textbf{62.1}        \\
			\hline
	\end{tabular}}
	\caption{Experiment results of ablation study on both ESC corpus and CTB corpus.}
	\label{Tab:Ablation}
\end{table*}

\par
We can first observe that our \textsf{DAPrompt} with all four PLMs has achieved better performance than the competitors. Even most of the competitors have used an advanced PLM like RoBERTa and BERT, to train an elaborate downstream task model or by adopting the prompt learning paradigm. This again validates the design objective of our deterministic assumption prompt learning, which pre-assumes the existence causal relation and next evaluates the assumption rationality, other than directly predicting the existence of causal relation between two events.

\par
We can also observe that using different PLMs do result in some performance variations. This is not unexpected. This might be attributed to that all the four PLMs employ a kind of Transformer-based model in pre-training on large-scale corpus, even each with some different training strategies. Finally, the \textsf{DAPrompt (DeBERTa)} has achieved the best performance, which applies a disentangled attention mechanism to encode context and position information separately. As such, we implement the remaining ablation experiments with DeBERTa.

\par
\textbf{Conventional Prompt Learning:} To compare our \textsf{DAPrompt} with conventional prompt model, we conduct experiments on a conventional prompt model with different prompt designs for ablation study.
\begin{itemize}
	\item \textsf{Prompt} is a conventional prompt model with discrete template and some answer words for prediction.
	\item \textsf{Prompt + Virtual Answer (VA)} uses virtual answer words in the conventional prompt model.
	\item \textsf{Prompt + Continuous Template (CT)} uses continuous template in the conventional prompt model.
	\item \textsf{Prompt + VA + CT} uses both virtual answer words and continuous template in the conventional prompt model
\end{itemize}

\par
The first group of Table~\ref{Tab:Ablation} presents the results using conventional prompt learning models. It is observed that the \textsf{Prompt} cannot outperform the \textsf{Prompt+VA} and \textsf{Prompt+CT} that use more representative virtual answer words for prediction and continuous template for automatically prompt template searching, respectively. The \textsf{Prompt+VA+CP} combining both the virtual answer and continuous template achieves better performance compare with the other conventional prompt models.

\par
Although these conventional prompt learning models have employed some advanced techniques, viz. the virtual answer and continuous template, they still cannot outperform our \textsf{DAPrompt} learning. This again validates our new design style of deterministic assumption first and rationality evaluation next, rather than the conventional style of predicting an answer word first and mapping it to some relation.

\par
\textbf{Module ablation study:} To examine the effectiveness of different modules in our \textsf{DAPrompt}, we design the following ablation study.
\begin{itemize}
	\item \textsf{DAPrompt w/ Single Mask (SiM)} uses one mask with an event mention to predict the other event for rationality evaluation.
	\item \textsf{DAPrompt w/ Shared MLM (ShM)} uses one MLM head for answer prediction of two masks.
	\item \textsf{DAPrompt w/ Event Tokens (ET)} uses the probability of predicted event mention for rationality evaluation.
\end{itemize}

\par
The second group of Table~\ref{Tab:Ablation} presents the results of ablation modules. We observe that none of them can outperform the full \textsf{DAPrompt} model.
This, however, is not unexpected. The \textsf{DAPrompt w/ Sim} misses one event's rationality for causality assumption evaluation; While the causal relation is between two events, thus both rationalities of two predicted events are useful for the assumption evaluation. The \textsf{DAPrompt w/ ShM} ignores the impact between two event predictions with one MLM classifier.

\par
Besides, the inferior performance of the \textsf{DAPrompt w/ ET} may be attributed to the large number of different event description words in the dataset, leading to an unbalance answer label set and inadequate training process. From our statistics, the 5,334 annotated event mentions in ESC corpus and 6,813 annotated event mentions in CTB corpus are described by totally 1,656 and 2,045 different words or phrases respectively, and some of them contain very few instances. On the other hand, this also validates our design of using virtual event tokens of \texttt{<E1>} and \texttt{<E2>} to for events' representations.

\section{Conclusion}
This paper has designed a novel style of prompt learning for event casualty identification, that is, first deterministic assumption and next rationality evaluation, with the considerations of how to best utilize the encyclopedia-like knowledge embedded in a language model. We first assume the existence of causal relation between events and design a deterministic assumption template concatenating with the input event pair to predict event' tokens. We next use the probabilities of correctly predicted input events to evaluate the assumption rationality for the final event causality decision. Experiments on the ESC corpus validate our design objective in terms of significant performance improvements over all competitors and achieving the new state-of-the-art performance.

\par
In our future work, we shall investigate the fusion of document-level semantics into the assumption evaluation for the ECI task, as well as identifying the direction of event causalities. We are also interested in investigating the applicability of such deterministic assumption prompt learning for other NLP tasks.

\clearpage
\section*{Acknowledgements}
This work is supported in part by National Natural Science Foundation of China (Grant No: 62172167). The computation is completed in the HPC Platform of Huazhong University of Science and Technology.

\bibliography{mybibfile}
\bibliographystyle{acl_natbib}

%

\end{document}